\documentclass[letterpaper]{article} 
\usepackage[]{aaai24}  
\usepackage{times}  
\usepackage{helvet}  
\usepackage{courier}  
\usepackage[hyphens]{url}  
\usepackage{graphicx} 
\usepackage{natbib}  
\usepackage{caption} 
\usepackage{algorithm}
\usepackage{algorithmic}
\usepackage{amsmath}
\usepackage{amsfonts}       
\usepackage{nicefrac}       
\usepackage{microtype}      
\usepackage{xcolor}         
\usepackage{booktabs}       
\usepackage{makecell}
\usepackage{multirow}
\graphicspath{{figs/}}
\usepackage{tabularx}
\usepackage{bbding}
\usepackage{pifont}
\usepackage{bm}

\newcommand{\op}[1]{\operatorname{#1}}

\newcommand{\cmark}{\ding{51}}%
\newcommand{\xmark}{\ding{55}}%

\usepackage{newfloat}
\usepackage{listings}
\DeclareCaptionStyle{ruled}{labelfont=normalfont,labelsep=colon,strut=off} 
\floatstyle{ruled}
\newfloat{listing}{tb}{lst}{}
\floatname{listing}{Listing}
\nocopyright 

\title{AAAI Press Anonymous Submission\\Instructions for Authors Using \LaTeX{}}
\author{
    Qingyu Wang\textsuperscript{\rm 1,2}\equalcontrib,
    Duzhen Zhang\textsuperscript{\rm 1,2}\equalcontrib,
    Tielin Zhang\textsuperscript{\rm 1,2}\footnote{Corresponding author.},
    Bo Xu\textsuperscript{\rm 1,2,3}\footnotemark[2]
}
\affiliations{
    \textsuperscript{\rm 1}School of Artificial Intelligence, University of Chinese Academy of Sciences, Beijing, China\\
    \textsuperscript{\rm 2}Research Center for Brain-inspired Intelligence, Institute of Automation, Chinese Academy of Sciences, Beijing, China\\
    \textsuperscript{\rm 3}Center for Excellence in Brain Science and Intelligence Technology, Chinese Academy of Sciences, Shanghai, China\\

    $\{$wangqingyu2022,zhangduzhen2019,tielin.zhang,xubo$\}$@ia.ac.cn
}

\title{Attention-free Spikformer: Mixing Spike Sequences with \\Simple Linear Transforms}
\usepackage{bibentry}

\begin{document}

\maketitle

\begin{abstract}
By integrating the self-attention capability and the biological properties of Spiking Neural Networks (SNNs), Spikformer applies the flourishing Transformer architecture to SNNs design. It introduces a Spiking Self-Attention (SSA) module to mix sparse visual features using spike-form Query, Key, and Value, resulting in the State-Of-The-Art (SOTA) performance on numerous datasets compared to previous SNN-like frameworks.
In this paper, we demonstrate that the Spikformer architecture can be accelerated by replacing the SSA with an unparameterized Linear Transform (LT) such as Fourier and Wavelet transforms. These transforms are utilized to mix spike sequences, reducing the quadratic time complexity to log-linear time complexity.
They alternate between the frequency and time domains to extract sparse visual features, showcasing powerful performance and efficiency.
We conduct extensive experiments on image classification using both neuromorphic and static datasets.
The results indicate that compared to the SOTA Spikformer with SSA, Spikformer with LT achieves higher Top-1 accuracy on neuromorphic datasets (\emph{i.e.}, CIFAR10-DVS and DVS128 Gesture) and comparable Top-1 accuracy on static datasets (\emph{i.e.}, CIFAR-10 and CIFAR-100).
Furthermore, Spikformer with LT achieves approximately $29$-$51\%$ improvement in training speed, $61$-$70\%$ improvement in inference speed, and reduces memory usage by $4$-$26\%$ due to not requiring learnable parameters.
\end{abstract}

\section{Introduction}

Spiking Neural Networks (SNNs), considered the third generation of neural networks~\cite{maass1997networks}, have garnered increasing attention due to their low power consumption, biological plausibility, and event-driven characteristics that align naturally with neuromorphic hardware~\cite{DBLP:conf/ijcai/ZhangZJW022,akopyan2015truenorth,davies2018loihi}.
By borrowing advanced architectures from Artificial Neural Networks (ANNs), SNNs have significantly improved their performance, including Spiking Recurrent Neural Networks~\cite{lotfi2020long}, ResNet-like SNNs~\cite{hu2021spiking,fang2021deep,zheng2021going,hu2021advancing}, and Spiking Graph Neural Networks~\cite{DBLP:conf/ijcai/Xu00LLP21,DBLP:conf/ijcai/ZhuPL0YL22}. 
The Transformer architecture~\cite{vaswani2017attention} has recently achieved rapid and widespread dominance in natural language processing~\cite{kenton2019bert,radford2018improving} and computer vision~\cite{dosovitskiyimage}. 
At the heart of the Transformer lies the self-attention mechanism, which connects each token/patch in the input sequence via a relevance-weighted basis of every other token/patch. 
Exploiting both the self-attention capability and the biological properties of SNNs, Spikformer~\cite{zhou2022spikformer} explores self-attention in SNNs for more advanced deep learning. 
It applies the flourishing Transformer architecture to SNN design and achieves State-Of-The-Art (SOTA) performance on numerous neuromorphic and static image classification datasets, surpassing previous directly-trained SNN-like frameworks.

To conform the properties of SNNs, Spikformer~\cite{zhou2022spikformer} introduces a spike-form self-attention variant called Spiking Self-Attention (SSA). In SSA, the Query, Key, and Value consist only of $0$ and $1$. The resulting attention map, computed from spike-form Query and Key, naturally exhibits non-negativeness. As a result, the attention map does not require the softmax operation for normalization, which leads to improved computational efficiency compared to Vanilla Self-Attention (VSA). However, SSA still exhibits an $\mathcal{O}(N^2)$ complexity, with $N$ representing the sequence length.

In this paper, our primary objective is to investigate whether more straightforward sequence mixing mechanisms can completely substitute the relatively complex SSA sub-layers in Spikformer architecture. 
To our surprise, we discover that a simple Linear Transform (LT), such as Fourier Transform (FT) or Wavelet Transform (WT), achieves comparable or even superior performance to SSA, despite having no learnable parameters. Moreover, LT exhibits higher computational efficiency with an excellent time complexity of $\mathcal{O}(N\log N)$.
LT utilizes fixed basis functions to extract features from the time or frequency domains, whereas SSA can be considered based on adaptive basis functions (\emph{i.e.}, attention weights). 
Although adaptive methods with learnable parameters are generally considered more flexible and perform better, they may not always be suitable in the spike context. 
This is because the attention weights calculated from spike-form Query and Key may lack precision for sequence mixing.
Consequently, the complex SSA mixing mechanism may not be necessary in this context, and an unparameterized LT mixing mechanism is more suitable for spike sequences with limited information. 

The primary contributions of this paper can be succinctly outlined as follows:

\begin{itemize}
    \item We demonstrate that even simple LTs, such as unparameterized FT and WT, are effective in extracting sparse visual features. The fact that such straightforward LTs yield promising results is surprising and suggests that SSA may not be the principal factor driving the Spikformer's performance.
    \item We introduce a novel Spikformer variant incorporating FT or WT for sequence mixing. Additionally, we comprehensively analyze the time complexity associated with different sequence mixing mechanisms, including SSA, FT, and WT.
    \item Extensive experiments validate that our proposed Spikformer with LT achieves superior Top-1 accuracy on neuromorphic datasets and comparable performance on static datasets compared to the original Spikformer with SSA. Moreover, our Spikformer with LT exhibits significantly enhanced computational efficiency, reducing memory usage by $4$-$26\%$ while achieving approximately $29$-$51\%$ and $61$-$70\%$ improvements in training and inference speeds.
\end{itemize}

\section{Related Work}

\subsection{Vision Transformers}
The transformer architecture, initially designed for natural language processing~\cite{vaswani2017attention}, has demonstrated remarkable success in various computer vision tasks, including image classification~\cite{dosovitskiyimage, yuan2021tokens}, semantic segmentation~\cite{wang2021pyramid, yuan2022volo}, object detection~\cite{carion2020end, zhudeformable, liu2021swin}, and low-level image processing~\cite{chen2021pre}. The critical component that contributes to the success of the transformer is the Self-Attention mechanism. In Vision Transformers (ViT), Self-Attention allows for capturing global dependencies and generating meaningful representations by weighting the features of image patches using dot-product operations between Query and Key, followed by softmax normalization~\cite{katharopoulos2020transformers, qincosformer}. The structure of ViT holds promise for a new class of SNNs, offering potential breakthroughs in overcoming performance limitations. Spikeformer~\cite{li2022spikeformer} merely replaces the activation function in the feedforward layers with spiking neuron models while retaining numerous non-spike operations. To conform to the calculation principles of SNNs, Spikformer~\cite{zhou2022spikformer} introduces an SSA module that utilizes spike-form Query, Key, and Value without employing softmax normalization. This development results in a powerful spiking ViT model for image classification, achieving SOTA performance on numerous datasets compared to previous directly-trained SNN-like frameworks.

\subsection{Spiking Neural Networks} 

In contrast to traditional ANNs that employ continuous floating-point values to convey information, SNNs utilize discrete spike sequences for computation and communication, offering a promising energy-efficient and biologically-plausible alternative~\cite{DBLP:conf/ijcai/ZhangZJW022}. 
The critical components of SNNs encompass spiking neuron models, optimization algorithms, and network architectures.

\paragraph{Spiking neuron models} serve as the fundamental processing units in SNNs, responsible for receiving continuous inputs and converting them into spike sequences. 
These models include Leaky Integrate-and-Fire (LIF)~\cite{dayan2005theoretical, wu2018spatio}, PLIF~\cite{fang2021incorporating}, Izhikevich~\cite{izhikevich2004spike,DBLP:conf/aaai/ZhangZJ022} neurons, among others.

\paragraph{Optimization algorithms} are primarily divided into two categories in deep SNNs: ANN-to-SNN conversion and direct training. 
In ANN-to-SNN conversion~\cite{cao2015spiking, hunsberger2015spiking, rueckauer2017conversion, DBLP:conf/iclr/BuFDDY022, meng2022training, wang2022signed}, a high-performance pre-trained ANN is converted into an SNN by replacing ReLU activation functions with spiking neurons. However, the converted SNN requires significant time steps to accurately approximate the ReLU activation, leading to substantial latency~\cite{han2020rmp}.
In direct training, SNNs are unfolded over simulation time steps and trained using backpropagation through time~\cite{lee2016training, shrestha2018slayer}. 
Since the event-triggered mechanism in spiking neurons is non-differentiable, surrogate gradients are employed to approximate the non-differentiable parts during backpropagation, using predefined gradient values~\cite{neftci2019surrogate, lee2020enabling}.

\paragraph{Network architectures}
With the advancements in ANNs, SNNs have improved their performance by incorporating advanced architectures from ANNs. These architectures include Spiking Recurrent Neural Networks~\cite{lotfi2020long}, ResNet-like SNNs~\cite{hu2021spiking, fang2021deep, zheng2021going, hu2021advancing}, and Spiking Graph Neural Networks~\cite{DBLP:conf/ijcai/Xu00LLP21, DBLP:conf/ijcai/ZhuPL0YL22}. However, exploring self-attention and Transformer models in the context of SNNs is relatively limited.
To address this, temporal attention has been proposed to reduce redundant simulation time step~\cite{yao2021temporal}. Additionally, an ANN-SNN conversion Transformer has been introduced, but it still retains VSA that do not align with the inherent properties of SNNs~\cite{mueller2021spiking}.
More recently, Spikformer~\cite{zhou2022spikformer} investigates the feasibility of implementing self-attention and Transformer models in SNNs using a direct training manner. Furthermore, some papers employ ANN-Transformers to process spike data, even though they may have ``Spiking Transformer" in their title~\cite{zhang2022spiking, zhang2022spike,DBLP:conf/aaai/WangZHWZX23}.

In this paper, we argue that the utilization of SSA is unnecessary in the context of SNNs.
We focus on exploring the potential of simpler sequence mixing mechanisms as viable substitutes for the relatively intricate SSA sub-layers present in Spikformer architecture. 
Furthermore, we present compelling evidence that simple LT, such as FT and WT, not only delivers comparable or even superior performance compared to SSA but also manifests enhanced computational efficiency, despite its lack of learnable parameters.

\section{Preliminaries}

\subsection{Spiking Neuron Model}

The spike neuron serves as the fundamental unit in SNNs. It receives the resultant current and accumulates membrane potential, which is subsequently compared to a threshold to determine whether a spike should be generated. In our paper, we consistently employ LIF at all Spike Neuron ($\mathcal{SN}$) layers.
The dynamic model of the LIF neuron is described as follows:
\begin{equation}
\begin{aligned}
H[t]&=V[t-1]+\frac{1}{\tau}\left(X[t]-\left(V[t-1]-V_{\text {reset }}\right)\right),\\
S[t]&=\mathcal{G}\left(H[t]-V_{t h}\right) , \\
V[t]&=H[t](1-S[t])+V_{\text {reset }} S[t],
\end{aligned}
\end{equation}
where $\tau$ represents the membrane time constant, and $X[t]$ denotes the input current at time step $t$. When the membrane potential $H[t]$ surpasses the firing threshold $V_{th}$, the spike neuron generates a spike $S[t]$. The Heaviside step function $\mathcal{G}(v)$ is defined as 1 when $v\geq 0$ and 0 otherwise. Following a spike event, the membrane potential $V[t]$ transitions to the reset potential $V_{reset}$ if a spike is produced, while it remains unchanged as $H[t]$ if no spike occurs.

\subsection{Spiking Self-Attention}
The Spikformer utilizes the SSA as its primary module for extracting sparse visual features and mixing spike sequences. 
Given input spike sequences denoted as $\bm{X} \in \mathbb{R}^{T \times N \times D}$, where $T$, $N$, and $D$ represent the time steps, sequence length, and feature dimension, respectively, SSA incorporates three key components: Query ($\bm{Q}$), Key ($\bm{K}$), and Value ($\bm{V}$). These components are initially obtained by applying learnable matrices $\bm{W}_Q, \bm{W}_K, \bm{W}_V \in\mathbb{R}^{D\times D }$ to the input sequences $\bm{X}$. Subsequently, they are transformed into spike sequences through distinct $\mathcal{SN}$ layers, formulated as:
\begin{equation}
\begin{aligned}
\bm{Q} &= {{\mathcal{SN}}_Q}(\op{BN}(\bm{X}\bm{W}_Q)),\\
\bm{K} &={{\mathcal{SN}_K}}(\op{BN}(\bm{X}\bm{W}_K)),\\
\bm{V} &= {{\mathcal{SN}_V}}(\op{BN}(\bm{X}\bm{W}_V)),
\label{eq:spikeqkv}
\end{aligned}
\end{equation}
where $\op{BN}$ denotes Batch Normalization operation and $\bm{Q},\bm{K},\bm{V} \in \mathbb{R}^{T \times N\times D}$. Inspired by VSA~\cite{vaswani2017attention}, SSA adds a scaling factor $s$ to control the large value of the matrix multiplication result, defined as:
\begin{equation}
\begin{aligned}
&{\rm{\text{SSA}}}(\bm{Q},\bm{K},\bm{V})={\mathcal{SN}}\left({\bm{Q}}~{\bm{K}^{\rm{T}}}~\bm{V} * s\right),\\
&\bm{X}^{\prime} ={\mathcal{SN}}(\op{BN}(\op{Dense}({\rm{\text{SSA}}}(\bm{Q},\bm{K},\bm{V})))),
\label{eq:ssa}
\end{aligned}
\end{equation}
where $\bm{X}^{\prime}\in\mathbb{R}^{T\times N \times D}$ are the updated spike sequences. It should be noted that SSA operates independently at each time step. In practice, $T$ represents an independent dimension for the $\mathcal{SN}$ layer. In other layers, it is merged with the batch size. Based on Equation (\ref{eq:spikeqkv}), the spike sequences $\bm{Q}$ and $\bm{K}$ produced by the $\mathcal{SN}$ layers $\mathcal{SN}_Q$ and $\mathcal{SN}_K$, respectively, naturally have non-negative values ($0$ or $1$). Consequently, the resulting attention map is also non-negative. Therefore, according to Equation (\ref{eq:ssa}), there is no need for softmax normalization to ensure the non-negativity of the attention map, and direct multiplication of $\bm{Q}$, $\bm{K}$, and $\bm{V}$ can be performed. This approach, compared to VSA, significantly improves computational efficiency.

However, it is essential to note that SSA remains an operation with a computational complexity of $\mathcal{O}(N^2)$.\footnote{Although SSA can be decomposed with an $\mathcal{O}(N)$ attention scaling, this complexity hides large constants, causing limited scalability in practical applications. For a more detailed analysis, refer to the \textbf{Time Complexity Anaysis of LT vs. SSA} Section.}
In the context of SNNs, we firmly believe that SSA is not essential, and there exist simpler sequence mixing mechanisms that can efficiently extract sparse visual features as alternatives.

\section{Method}
\begin{figure*}[tbp]
\centering
\includegraphics[width=0.94\textwidth]{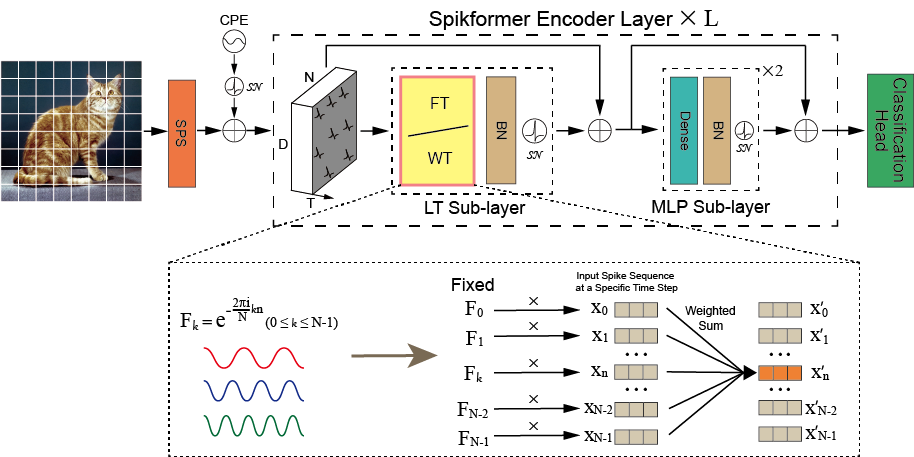}
\caption{The overview of our Spikformer with Linear Transforms (LT), which consists of a Spiking Patch Splitting (SPS) module with a Conditional Position Embedding (CPE) generator, multiple Spikformer encoder layers, and a classification head. Each Spikformer encoder layer consists of a LT sub-layer and a Multi-Layer Perceptron (MLP) sub-layer, both equipped with Batch Normalization (BN) and Spike Neuron ($\mathcal{SN}$) layers. Unlike original spiking self-attention sub-layer, our LT sub-layer relies on unparameterized fixed basis functions to mix an input spike sequence, can be implemented using either Fourier Transform (FT) or Wavelet Transform (WT) methods. Here, we illustrate with a discrete FT implementation along the sequence dimension. For simplicity, we focus solely on the operation process on a feature $x_n$ in the input spike sequence at a specific time step.}
\label{model}
\end{figure*}

Following a standard ViT architecture, Spikformer~\cite{zhou2022spikformer} incorporates several key components: the Spiking Patch Splitting (SPS) module, multiple Spikformer encoder layers, and a classification head for image classification tasks. 
The Spikformer encoder layer comprises a SSA sub-layer and a Multi-Layer Perception (MLP) sub-layer. 
To enhance the efficiency of the original Spikformer architecture, we substitute the SSA sub-layer with a LT sub-layer to efficiently mix spike sequences. 
In the following sections, we provide an overview of our Spikformer with LT (shown in Figure~\ref{model}), followed by a detailed explanation of the LT sub-layer. Lastly, we analyze the time complexity of LT versus SSA.

\subsection{Overall Architecture}
Figure~\ref{model} provides an overview of our Spikformer with LT. For a given 2D image sequence $\bm{I}\in \mathbb{R}^{T \times C\times H\times W}$,\footnote{In the neuromorphic dataset, the data shape is $\bm{I} \in \mathbb{R}^{T \times C\times H\times W}$, where $T$, $C$, $H$, and $W$ denote the time step, channel, height, and width, respectively. In static datasets, a 2D image $\bm{I}_s \in \mathbb{R}^{C\times H\times W}$ needs to be repeated $T$ times to form an image sequence.} the goal of the SPS module is to linearly project it into a $D$-dimensional spike-form feature and split this feature into a sequence of $N$ flattened spike-form patches $\bm{P}\in \mathbb{R}^{T \times N \times D}$. 
Following the approach of the original Spikformer~\cite{zhou2022spikformer}, the SPS module employs convolution operations to introduce inductive bias~\cite{xiao2021early,hassani2021escaping}.
To generate spike-form relative position embedding (RPE), we utilize the Conditional Position Embedding (CPE) generator~\cite{chu2021twins} in the same manner as the original Spikformer. This RPE is then added to the patch sequence $\bm{P}$, resulting in $\bm{X}_0\in \mathbb{R}^{T \times N \times D}$. 
The $\bm{X}_0$ is subsequently passed through the $L$-layer Spikformer encoder. Unlike the original Spikformer with SSA~\cite{zhou2022spikformer}, our Spikformer encoder layer consists of a LT sub-layer and an MLP sub-layer, both with BN and $\mathcal{SN}$ layers. Residual connections are applied to both the LT and MLP sub-layers.
The LT sub-layer serves as a critical component in our Spikformer encoder layer, providing an efficient method for mixing spike sequences. 
We have provided two implementations for LT, namely FT, and WT, which alternate between the frequency and time domains to extract sparse visual features. These implementations will be thoroughly analyzed in the \textbf{Linear Transform} Section.
Following the processing in the Spikformer encoder, a Global Average-Pooling (GAP) operation is applied to the resulting spike features, generating a $D$-dimensional feature. This feature is then fed into the fully connected-layer Classification Head (CH), which produces the prediction $\bm{Y}$. The formulation of our Spikformer with LT can be expressed as follows:
\begin{align}
\bm{P}&={\rm{SPS}}\left(\bm{I}\right),  \\
{\bm{\text{RPE}}}&={\rm{CPE}}(\bm{P}), \\
\bm{X}_0 &= \bm{P} + \bm{\text{RPE}}, \\
\bm{X}^{\prime}_l &= \mathcal{SN}(\rm{BN}({\rm{LT}}(\bm{X}_{l-1}))) + \bm{X}_{l-1}, \\
 \bm{X}_l &= \mathcal{SN}(\rm{BN}({\rm{MLP}}(\bm{X}^{\prime}_l))) + \bm{X}^{\prime}_l, \\
 \bm{Y} &= \op{CH}(\op{GAP}(\bm{X}_L)),
\end{align}
where ${{\bm{I}}} \in \mathbb{R}^{T \times C\times H\times W}$, $\bm{P}\in \mathbb{R}^{T\times N\times D}$, ${\bm{\text{RPE}}}\in \mathbb{R}^{T \times N\times D}$, $\bm{X}_0\in \mathbb{R}^{T \times N\times D}$, $\bm{X}^{\prime}_l\in \mathbb{R}^{T \times N\times D}$, $\bm{X}_l\in \mathbb{R}^{T \times N\times D}$, and $l=1...L$.

\subsection{Linear Transform}\label{sec:lt}
Given input spike sequences $\bm{X} \in \mathbb{R}^{T \times N \times D}$, the LT sub-layer alternates between the frequency and time domains to extract sparse visual features. These features are then transformed into spiking sequences $\bm{X}^{\prime} \in \mathbb{R}^{T \times N \times D}$ through a $\mathcal{SN}$ layer. The formulation can be expressed as follows:
\begin{align}
      \bm{X}^{\prime} = \mathcal{SN}(\rm{BN}(\rm{LT}(\bm{X}))),
\end{align}
In contrast to the SSA sub-layer in Equation~(\ref{eq:ssa}), the LT sub-layer does not involve any learnable parameters or Self-Attention calculations. Next, we have introduced two implementations of LT: FT and WT.

\paragraph{Fourier Transform} The FT decomposes a function into its constituent frequencies~\cite{gonzales1987digital}. For an input spike sequence $\bm{x} \in \mathbb{R}^{N \times D}$ at a specific time step in $\bm{X}$, we utilize the FT to mix information from different dimensions, including 1D-FT and 2D-FT.

The discrete 1D-FT along the sequence dimension $\mathcal{F}_{\rm{seq}}$ to extract sparse visual features is defined by the equation:
\begin{equation}
    \label{eq:1d-dft}
    \bm{x}^{\prime}_n =\mathcal{F}_{\rm{seq}}(\bm{x}_n) =\sum_{k=0}^{N-1} \bm{x}_k e^{-{\frac{2\pi i}{N}} k n}, n=0,...,N-1, 
\end{equation}
where $i$ represents the imaginary unit. For each value of $n$, the discrete 1D-FT generates a new representation $\bm{x}^{\prime}_n\in\mathbb{R}^D$ as a sum of all of the original input spike features $\bm{x}_n\in\mathbb{R}^D$.\footnote{It is important to note that the weights in Equation~(\ref{eq:1d-dft}) are fixed constant and can be pre-calculated for all spike sequences.} 

Similarly, the discrete 2D-FT along the feature and sequence dimensions $\mathcal{F}_{\rm{seq}}(\mathcal{F}_{\rm{f}})$ is defined by the equation:
\begin{equation}
    \label{eq:2d-dft}
    \bm{x}^{\prime}_n =\mathcal{F}_{\rm{seq}}(\mathcal{F}_{\rm{f}}(\bm{x}_n)), n=0,...,N-1, 
\end{equation}
Equations (\ref{eq:1d-dft}) and (\ref{eq:2d-dft}) only consider the real part of the result. Therefore, there is no need to modify the subsequent MLP sub-layer or output layer to handle complex numbers.

\paragraph{Wavelet Transform} 
WT is developed based on FT to overcome the limitation of FT in capturing local features in the time domain~\cite{gonzales1987digital}. 

The discrete 1D-WT along the sequence dimension $\mathcal{W}_{\rm{seq}}$ to extract sparse visual features is defined by the equation:
\begin{equation}
\begin{aligned}
    \label{eq:1d-dwt}
    &\bm{x}^{\prime}_{n} =\mathcal{W}_{\rm{seq}}(\bm{x}_{n})=
    \frac{1}{\sqrt{N}}\Big[\bm{T}_{\varphi}(0,0)*\varphi(\bm{x}_{n}) \\
    & + \sum\limits_{j=0}^{J-1}\sum\limits_{k=0}^{2^j-1}\bm{T}_{\psi}(j,k)*\psi_{j,k}(\bm{x}_{n})\Big],\ \ \text{where}\\
    &\bm{T}_{\varphi}(0,0) = \frac{1}{\sqrt{N}}\sum\limits_{k=0}^{N-1}\bm{x}_{k}*\varphi({\bm{x}_{k}}),\\
&\bm{T}_{\psi}(j,k)=\frac{1}{\sqrt{N}}\sum\limits_{k^{\prime}=0}^{N-1}\bm{x}_{k^{\prime}}*\psi_{j,k}({\bm{x}_{k^{\prime}}}),
\end{aligned}
\end{equation}
where $n=0,...,N-1$, $N=2^J$ ($N$ is typically a power of $2$), $*$ denotes element-wise multiplication, $\bm{T}_{\varphi}(0,0)$ are the approximation coefficients, $\bm{T}_{\psi}(j,k)$ are the detail coefficients, $\varphi(x)$ is the scaling function, and $\psi_{j,k}(x)=2^{j/2}\psi(2^jx-k)$ is the wavelet function. 
Here, we use the Haar scaling function and Haar wavelet function, defined by the equation:
\begin{align}
\varphi(x)&=\left\{
	\begin{aligned}
	&1 \quad 0\leq x<1\\
	&0 \quad \rm{otherwise}\\
	\end{aligned}
	\right.,\\
 \psi(x)&=\left\{
	\begin{aligned}
	&1 \quad 0\leq x<0.5\\
 	&-1 \quad 0.5\leq x<1\\
	&0 \quad \rm{otherwise}\\
	\end{aligned}
	\right.,
\end{align}

In the subsequent experimental section, we also delve into the exploration of different basis functions as well as their potential combinations.

Similarly, the discrete 2D-WT along the feature and sequence dimensions $\mathcal{W}_{\rm{seq}}(\mathcal{W}_{\rm{f}})$ is defined by the equation:
\begin{equation}
    \label{eq:2d-dwt}
    \bm{x}^{\prime}_n =\mathcal{W}_{\rm{seq}}(\mathcal{W}_{\rm{f}}(\bm{x}_n)), n=0,...,N-1, 
\end{equation}

\subsection{Time Complexity Anaysis of LT vs. SSA}\label{complex}

We conduct a time complexity analysis of different mixing mechanisms, including SSA, FT, and WT. The results of the time complexity analysis are presented in Table~\ref{tab:complex}. In the subsequent experimental section, we also conduct a more specific comparison of the training and inference speeds of different mixing mechanisms under the same conditions.

In SSA (Equation (\ref{eq:ssa})), since there is no softmax operation, the order of calculation between $\bm{Q}$, $\bm{K}$, and $\bm{V}$ can be changed: either $\bm{QK}^{\rm{T}}$ followed by $\bm{V}$, or $\bm{K}^{\rm{T}}\bm{V}$ followed by $\bm{Q}$. The former has a time complexity of $\mathcal{O}(N^2d)$, while the latter has a time complexity of $\mathcal{O}(Nd^2)$, where $d$ is the feature dimension per head.\footnote{In practice, the SSA in Equation (\ref{eq:ssa}) can be extended to multi-head SSA. In this case, $d = D/H$, where $H$ is the number of heads.} The second complexity, $\mathcal{O}(Nd^2)$, cannot be simply considered as $\mathcal{O}(N)$ due to the large constant $d^2$ involved. 
Only when the sequence length $N$ is significantly larger than the feature dimension per head $d$ does it demonstrate a significant computational efficiency advantage over the first complexity, $\mathcal{O}(N^2d)$.

In our implementation, we utilize the Fast Fourier Transform (FFT) algorithm to compute the discrete FT.
Specifically, we employ the Cooley-Tukey algorithm~\cite{cooley1965algorithm,frigo2005design}, which recursively expresses the discrete FT of a sequence of length $N=N_1N_2$ in terms of $N_1$ smaller discrete FTs of size $N_2$, reducing the time complexity to $\mathcal{O}(N\log N)$ for discrete 1D-FT along the sequence dimension. Similarly, for discrete 2D-FT first along the feature dimension and then along the sequence dimension, the time complexity is $\mathcal{O}(D\log D)+\mathcal{O}(N\log N)$. In general, the complexity of WT is comparable to that of FFT~\cite{gonzales1987digital}.

\begin{table}[t]
    \centering

    \resizebox{1.0\columnwidth}{!}{
    \begin{tabular}{c | c }
        \hline
        Mixing Mechanisms & Time Complexity \\ \hline\hline
        SSA & $\mathcal{O}(N^2d)$\ \rm{or}\ $\mathcal{O}(Nd^2)$\\
        1D-FFT & $\mathcal{O}(N\log N)$\\ 
        2D-FFT & $\mathcal{O}(D\log D)+\mathcal{O}(N\log N)$ \\ 
        2D-WT & $\mathcal{O}(D\log D)+\mathcal{O}(N\log N)$ \\ \hline
    \end{tabular} 
    }
 \caption{The time complexity results for different mixing mechanisms. Generally, we have $N = 64$, $D = 384\ \text{or}\ 256$, and $d = 32$. The logarithms mentioned are in base $2$.}
        \label{tab:complex}

\end{table}

\begin{table*}[t]
  \centering

\resizebox{1.0\textwidth}{!}{
    \begin{tabular}{cccccccc}
    \toprule
    Methods & Architecture & Spikes& \makecell{Time Step\\(DVS10\textbf{/}128)} & \makecell{Memory\\(DVS10\textbf{/}128)\\(GB)}  & \makecell{Training Speed\\(DVS10\textbf{/}128)\\(ms/batch)}& \makecell{Inference Speed\\(DVS10\textbf{/}128)\\(ms/batch)} & \makecell{Top-1 acc.\\(DVS10\textbf{/}128)} \\
   \midrule
   LIAF-Net~\cite{wu2021liaf} & -- & \xmark & 10\textbf{/}60 & -- & -- & --& 70.4\textbf{/} 97.6\\
    TA-SNN~\cite{yao2021temporal} & -- & \xmark & 10\textbf{/}60 & -- & -- & --& 72.0\textbf{/} 98.6\\
   \midrule
     Rollout~\cite{kugele2020efficient} & -- & \cmark & 48\textbf{/}240 & -- & -- & --& 66.8\textbf{/} 97.2\\
    DECOLLE~\cite{kaiser2020synaptic} & -- & \cmark & --\textbf{/}500 & -- & -- & --& --\textbf{/} 95.5\\
    tdBN~\cite{zheng2021going} & ResNet-19 & \cmark & 10\textbf{/}40 & -- & -- & --& 67.8\textbf{/} 96.9\\

    PLIF~\cite{fang2021incorporating} & -- & \cmark & 20\textbf{/}20 & -- & -- & --& 74.8\textbf{/} 97.6\\
    SEW-ResNet~\cite{fang2021deep} & Wide-7B-Net & \cmark & 16\textbf{/}16 & -- & -- & --& 74.4\textbf{/} 97.9\\

    Dspike~\cite{li2021differentiable} & -- & \cmark & 10\textbf{/}-- & -- & -- & --& $\rm{75.4}^{\dagger}$\textbf{/}--\\

    SALT~\cite{kim2021optimizing} & -- & \cmark & 20\textbf{/}-- & -- & -- & --& 67.1\textbf{/}--\\

    DSR~\cite{meng2022training} & -- & \cmark & 10\textbf{/}-- & -- & -- & --& $\rm{77.3}^{\dagger}$\textbf{/}--\\
    
    \midrule
    \multicolumn{1}{c}{\multirow{2}{*}{SSA~\cite{zhou2022spikformer}}} 
                                                  
    & {Spikformer-2-256} & \cmark& 16\textbf{/}16 & -- & --  & -- &  $\rm{80.9}^{\dagger}$\textbf{/}{98.3}\\
    & {Spikformer-2-256}* & \cmark& 16\textbf{/}16 & 9.02\textbf{/}9.03 & 76\textbf{/}246  & 30\textbf{/}105 &  $\rm{79.7}^{\dagger}$\textbf{/}{98.2}\\
    \midrule
                                                  
      \textbf{1D-FFT (Our LT)} & {Spikformer-2-256} & \cmark& 16\textbf{/}16 & 8.67\textbf{/}\textbf{8.71} & \textbf{51}\textbf{/}\textbf{121} & \textbf{11}\textbf{/}\textbf{32} &  $\rm{81.1}^{\dagger}$\textbf{/}{\textbf{99.1}}\\
      \textbf{2D-FFT (Our LT)} & {Spikformer-2-256} & \cmark& 16\textbf{/}16 & \textbf{8.54}\textbf{/}8.74 & 55\textbf{/}135  & 21\textbf{/}37 &  $\rm{81.2}^{\dagger}$\textbf{/}{98.4}\\
      \textbf{2D-WT-Haar (Our LT)} & {Spikformer-2-256} & \cmark& 16\textbf{/}16 & 8.70\textbf{/}8.73 & 62\textbf{/}139  & 23\textbf{/}46 &  $\rm{\textbf{81.6}}^{\dagger}$\textbf{/}{98.5}\\

    \bottomrule
    \end{tabular}%
}
\caption{Performance comparison of our method with existing methods on CIFAR10-DVS and DVS128 Gesture (DVS10/128). Our Spikformer with LT outperforms Spikformer with SSA on both datasets in terms of Top-1 acc. Notably, our method also exhibits a significantly smaller memory footprint and faster training and inference speeds. The symbol Spikes=\xmark\ denotes the use of floating-point spikes, and text in \textbf{bold} indicates the best results. $*$ means our reproduced results using open codebases, ensuring a fair comparison of memory, training, and inference speeds under identical operating conditions. Other baseline results are directly cited from the original Spikformer~\cite{zhou2022spikformer}. $\dagger$ indicates the inclusion of data augmentation. It is worth mentioning that Spikformer-2-256 signifies a configuration with $2$ Spikformer encoder layers and a feature dimension of $256$.}
  \label{tab:dvs}%
\end{table*}%

\begin{table*}[t]
  \centering

\resizebox{1.0\textwidth}{!}{
    \begin{tabular}{cccccccc}
    \toprule
    Methods & Architecture& \makecell{Time\\Step} & \makecell{Param\\(M)}& \makecell{Memory\\(GB)}  & \makecell{Training Speed\\(ms/batch)}& \makecell{Inference Speed\\(ms/batch)} & \makecell{Top-1 acc.\\(CIFAR-10\textbf{/}100)} \\
   \midrule
    Hybrid training~\cite{rathienabling} &VGG-11 &125& 9.27&--&--&--&92.22\textbf{/}67.87\\
    Diet-SNN~\cite{rathi2020diet} &ResNet-20 &10\textbf{/}5 &0.27 &--&--&--& 92.54\textbf{/}64.07\\
    STBP~\cite{wu2018spatio} &CIFARNet &12&17.54&--&--&--& 89.83\textbf{/}--\\
   STBP NeuNorm~\cite{wu2019direct} &CIFARNet &12&17.54 &--&--&-- &90.53\textbf{/}--\\
    TSSL-BP~\cite{zhang2020temporal} &CIFARNet &5 &17.54 &--&--&--& 91.41\textbf{/}--\\
    STBP-tdBN~\cite{zheng2021going} &ResNet-19 & 4 &12.63  &--&--&--&  92.92\textbf{/}70.86\\
   TET~\cite{dengtemporal}  &ResNet-19  & 4 &12.63 &--&--&--&  94.44\textbf{/}74.47\\

   \midrule
   \multicolumn{1}{c}{\multirow{2}{*}{{ANN}}}  &ResNet-19 &1 &12.63  &--&--&--&   94.97\textbf{/}75.35\\
     & {Transformer-4-384}  &1 & {9.32}  &--&--&--&  \textbf{96.73}\textbf{/}\textbf{81.02} \\
    \midrule
    \multicolumn{1}{c}{\multirow{2}{*}{SSA~\cite{zhou2022spikformer}}} 
                                                  
    & {Spikformer-4-384}& 4 &9.32   &--&--&--&  \textbf{95.51}\textbf{/}\textbf{78.21}\\
    & {Spikformer-4-384}* & 4 &9.32   &11.69&166&31&  {95.3}\textbf{/}{77.7}\\
    \midrule

  \textbf{1D-FFT (Our LT)} & {Spikformer-4-384}& 4 &\textbf{6.96}    &\textbf{8.61}&\textbf{118}&\textbf{12}&  {94.9}\textbf{/}{77.0}\\
    \textbf{2D-FFT (Our LT)}  & {Spikformer-4-384}& 4 &\textbf{6.96}    &8.75&122&13&  {95.1}\textbf{/}{77.3}\\
\textbf{2D-WT-Haar (Our LT)}  & {Spikformer-4-384}& 4 &\textbf{6.96}    &9.33&121&19&  {95.2}\textbf{/}{77.1}\\
    \bottomrule
    \end{tabular}%
    }
    \caption{Performance comparison of our method with existing methods on CIFAR-10/100.
Our Spikformer with LT achieves comparable Top-1 acc. on both datasets when compared to Spikformer with SSA. Importantly, our method demonstrates a significantly smaller memory footprint and faster training and inference speeds. The \textbf{bold} font denotes the highest result. $*$ represents our reproduced results using open codebases, ensuring a fair comparison of memory, training, and inference speeds under equal operating conditions. Other baseline results are directly cited from the original Spikformer~\cite{zhou2022spikformer}. Spikformer-4-384 refers to Spikformer with $4$ encoder layers and $384$ feature dimensions.}
  \label{tab:cifar}%
\end{table*}%

\section{Experiments}

We conduct experiments on neuromorphic datasets CIFAR10-DVS and DVS128 Gesture~\cite{amir2017low}, as well as static datasets CIFAR-10 and CIFAR-100 to evaluate the performance of Spikformer with LT. The Spikformer with LT is trained from scratch and compared against SOTA Spikeformer with SSA and previous SNN-like frameworks. More analysis is also conducted to analyze the effects of different WT basis functions and their combinations.

\subsection{Experimental Settings}
To ensure fair comparisons, we adhere to the configurations of SOTA Spikeformer with SSA~\cite{zhou2022spikformer} for datasets, implementation details, evaluation metrics, and baselines.

\paragraph{Neuromorphic Datasets} 

CIFAR10-DVS is a neuromorphic dataset created by shifting image samples to simulate their capture by a DVS camera. It offers $9,000$ training samples and $1,000$ test samples. DVS128 Gesture, on the other hand, is a gesture recognition dataset featuring $11$ hand gesture categories from $29$ individuals captured under $3$ different illumination conditions~\cite{amir2017low}.

For both datasets, CIFAR10-DVS and DVS128 Gesture, where the image size is $128 \times 128$, we employ the SPS module with a patch size of $16 \times 16$. This module splits each image into a sequence with length $N$ of $64$ and feature dimension $D$ of $256$.
We utilize $2$ Spikformer encoder layers and set the time-step of the spiking neuron to $16$. The training process consists of $106$ epochs for CIFAR10-DVS and $200$ epochs for DVS128 Gesture. We employ the AdamW optimizer with a batch size of $16$. The learning rate is initialized to $0.1$ and reduced using cosine decay. Additionally, data augmentation techniques, as described in~\cite{li2022neuromorphic}, are applied specifically to the CIFAR10-DVS dataset.

\paragraph{Static Datasets}

CIFAR-10/100 dataset consists of $50,000$ training images and $10,000$ test images, each with a resolution of $32\times 32$. 
To process these images, we employ the SPS module with a patch size of $4 \times 4$, which splits each image into a sequence of length $N=64$ and a feature dimension of $D=384$.
For the Spikformer encoder layers, we use $4$ layers, and the time-step of the spiking neuron is set to $4$. 
During training, we utilize the AdamW optimizer with a batch size of $128$. The training process spans $400$ epochs, with a cosine-decay learning rate starting at $0.0005$.
Following the approach outlined in~\cite{yuan2021tokens}, we apply standard data augmentation techniques such as random augmentation, mixup, and cutmix during training.

\begin{table}[t]
\setlength{\tabcolsep}{3pt}
\centering

\resizebox{1.0\columnwidth}{!}{
    \begin{tabular}{ccc}
    \toprule
    Methods & Architecture  & \makecell{DVS10/128\\Top-1 acc.} \\
   \midrule
       \multicolumn{1}{c}{\multirow{2}{*}{SSA~\cite{zhou2022spikformer}}}                    
    & {Spikformer-2-256}  & {80.9}\textbf{/}{98.3}\\
    & {Spikformer-2-256}* &  79.7\textbf{/}{98.2}\\
    \midrule
    2D-WT-Db1 & {Spikformer-2-256} &  81.4\textbf{/}{\textbf{98.7}}\\
2D-WT-Bior1.1  & {Spikformer-2-256} &  80.2\textbf{/}{98.2}\\
 2D-WT-Rbio1.1  & {Spikformer-2-256} &  80.4\textbf{/}{98.1}\\
 2D-WT-Combination & {Spikformer-2-256} &  80.4\textbf{/}{98.2}\\
2D-WT-Haar & {Spikformer-2-256} &  \textbf{81.6}\textbf{/}{98.5}\\

    \bottomrule
    \end{tabular}%
    }
    \caption{Analysis results about different WT basis functions on CIFAR10-DVS and DVS128 Gesture (DVS10/128). The bold font means the best. $*$ represents our reproduced results with the open codebases.}
  \label{tab:ablation}%
\vspace{+1.5mm}
\end{table}%
\vspace{+1.5mm}

We evaluate the performance of the image classification task using Top-1 accuracy (Top-1 acc.) as the performance metric. Furthermore, we compare the memory usage, training speed, and inference speed of the original Spikformer with SSA and our Spikformer with LT under identical operating conditions.
The training speed refers to the time required for the forward and back-propagation of a batch of data, while the inference speed denotes the time taken for the forward-propagation of a batch of data in milliseconds (ms). To minimize variance, we calculate the average time spent over 100 batches.
To conduct the experiments, we implement the models using PyTorch~\cite{paszke2019pytorch}, SpikingJelly~\cite{SpikingJelly}\footnote{\url{https://github.com/fangwei123456/spikingjelly}}, and the PyTorch image models library (Timm)\footnote{\url{https://github.com/rwightman/pytorch-image-models}}.
All experiments are conducted on an AMD EPYC $7742$ server equipped with $3$ NVIDIA A100 GPUs, each with $40$GB of memory.

\subsection{Main Experimental Results}\label{results}
The performance comparison of our Spikformer with LT, SOTA Spikformer with SSA, and previous SNN-like methods on neuromorphic datasets is presented in Table~\ref{tab:dvs}.
In the case of CIFAR10-DVS, our Spikformer with 2D-WT-Haar achieves an accuracy of $81.6\%$, surpassing the SOTA Spikformer with SSA (Original result: $80.9\%$, Our reproduction: $79.7\%$).
Our other LT variants also deliver comparable or better accuracy.
Under identical hyper-parameter configuration and operating conditions, our LT achieves approximately $5\%$ memory savings as it does not require learnable parameters. Additionally, it improves training speed and inference speed by approximately $33\%$ and $63\%$, respectively, compared to SSA. On DVS128 Gesture, our Spikformer with 1D-FFT achieves a $0.8\%$ higher accuracy than Spikformer with SSA, also outperforming TA-SNN~\cite{yao2021temporal} which utilizes floating-point spikes for forward propagation.
Similarly, our other LT variants achieve comparable or better accuracy.
Furthermore, under the same hyper-parameter configuration and operating conditions, our LT achieves approximately $4\%$ memory savings and enhances training speed and inference speed by approximately $51\%$ and $70\%$, respectively, compared to SSA.

Table~\ref{tab:cifar} presents a performance comparison of our Spikformer with LT, SOTA Spikformer with SSA, and other SNN-like methods on the CIFAR-10 and CIFAR-100 static datasets.
Across both datasets, our LT variants demonstrate comparable Top-1 accuracy compared to Spikformer with SSA.
Notably, our Spikformer with LT achieves several advantages under the same hyper-parameter configuration and operating conditions. It reduces the number of parameters by approximately $25\%$, resulting in memory savings of around $26\%$. Additionally, it enhances training and inference speed by approximately $29\%$ and $61\%$, respectively, compared to Spikformer with SSA.

\subsection{More Analysis}\label{more_analysis}

In our main experiments, the Haar basis function is used as the default choice for our WT. However, we also conducted experiments with other basis functions and their combinations, presented in Table~\ref{tab:ablation}.

Interestingly, most alternative basis functions and their combinations yield similar Top-1 accuracy on the CIFAR10-DVS and DVS128 Gesture datasets. In fact, their performance is comparable to or even better than that of Spikformer with SSA.
It is worth noting that the limited selection of WT basis functions employed in this study has shown minimal impact on classification accuracy. However, it is essential to highlight that WT offers numerous different basis function options, which we have not explored comprehensively due to time constraints. Investigating the influence of more basis function choices on the accuracy, as well as the possibility of identifying superior basis functions, is an avenue for future research.

\section{Conclusion}

This paper introduces several substantial contributions through an in-depth exploration of simplified sequence mixing mechanisms for Spikformer. Firstly, our study demonstrates the robust capacity of a straightforward LT mechanism to effectively model diverse visual dependencies present in images. This not only underscores the potency of simplicity in spiking sequence mixing but also provides a fundamental basis for subsequent advancements.
Building upon this insight, we put forward the concept of Spikformer with LT. In this paradigm, we replace the conventional parameterized SSA with an unparameterized FT or WT. By adopting these unparameterized approaches, we aim to streamline the model while retaining or even enhancing its performance in spiking sequence processing tasks.
The culmination of these efforts is a Spikformer with LT that not only refines the model's architecture but also yields substantial practical advantages. 
Notably, our Spikformer with LT results in memory usage reductions ranging from 4\% to 26\%. Simultaneously, it significantly accelerates both training speed, with improvements of approximately 29\% to 51\%, and inference speed, showcasing enhancements of 61\% to 70\%, when compared to the original Spikformer equipped with SSA. 
Moreover, it achieves better accuracy on neuromorphic datasets (\emph{i.e.}, CIFAR10-DVS and DVS128 Gesture) and demonstrates only nominal accuracy costs on static datasets (\emph{i.e.}, CIFAR-10 and CIFAR-100).

\newpage

\bibliography{aaai24}

\end{document}